\documentclass[journal]{IEEEtran}
\ifCLASSINFOpdf
\else
\fi
\usepackage{float}
\usepackage{epsfig}
\usepackage{graphicx}
\usepackage{amsmath}
\usepackage{amssymb}
\usepackage{flushend,epsfig}
\usepackage{booktabs}
\usepackage{lineno,hyperref}
\usepackage{multirow}
\usepackage{color}

\begin{document}
%
\title{Pose-adaptive Hierarchical Attention Network 
 for Facial Expression Recognition}
%
%
%

\author{Yuanyuan~Liu,~\IEEEmembership{Member,~IEEE}, Jiyao~Peng, Jiabei~Zeng,~\IEEEmembership{Member,~IEEE}, and Shiguang Shan\textsuperscript{*}, {Member,~IEEE}
\thanks{This work was supported by the National Natural Science Foundation of China (No.61602429). Corresponding author: Shiguang Shan  (e-mail: sgshan@ict.ac.cn).}
\thanks{Y. Liu and Jiyao Peng are with the School of Geography and Information Engineering, China University of Geosciences, Wuhan, 430074, China (e-mail: liuyy@cug.edu.cn;).

J. Zeng is with the Key Laboratory of Intelligent Information Processing, Institute of Computing Technology, Chinese Academy of Sciences, Beijing 100190, China (e-mail: jiabei.zeng@vipl.ict.ac.cn).

S. Shan is with the Key Laboratory of Intelligent Information Processing, Center for Excellence in Brain Science and Intelligence Technology, Institute of Computing Technology, Chinese Academy of Sciences, Beijing 100190, China, and also with the University of Chinese Academy of Sciences, Beijing 100049, China (e-mail: sgshan@ict.ac.cn).
}
\thanks{Manuscript received April X, 2019; revised August X, 2019.}}

%
%

\markboth{Journal of \LaTeX\ Class Files,~Vol.~14, No.~8, August~2015}%
{Shell \MakeLowercase{\textit{et al.}}: Bare Demo of IEEEtran.cls for IEEE Journals}
%



\maketitle
\begin{abstract}
Multi-view facial expression recognition (FER) is a challenging task because the appearance of an expression varies in poses. 
To alleviate the influences of poses, recent methods either perform pose normalization or learn separate FER classifiers for each pose.
However, these methods usually have two stages and rely on good performance of pose estimators.
Different from existing methods, we propose a pose-adaptive hierarchical attention network (PhaNet) that can jointly recognize the facial expressions and poses in unconstrained environment.
Specifically, PhaNet discovers the most relevant regions to the facial expression by an attention mechanism in hierarchical scales, and the most informative scales are then selected to learn the pose-invariant and expression-discriminative representations. 
PhaNet is end-to-end trainable by minimizing the hierarchical attention losses, the FER loss and pose loss with dynamically learned loss weights.  
We validate the effectiveness of the proposed PhaNet on three multi-view datasets (BU-3DFE, Multi-pie, and KDEF) and two in-the-wild FER datasets (AffectNet and SFEW). Extensive experiments demonstrate that our framework outperforms the state-of-the-arts under both within-dataset and cross-dataset settings, achieving the average accuracies of 84.92\%, 93.53\%, 88.5\%, 54.82\% and 31.25\% respectively. 
\end{abstract}


%
\IEEEpeerreviewmaketitle

\section{Introduction}

\IEEEPARstart{F}{acial} expressions convey cues about the emotional state of human beings and they serve as import affect signals. Hence, facial expression recognition (FER) has become a hot research topic of human-computer interaction. Automated FER is crucial to applications such as digital entertainment, customer service, driver monitoring, emotion robot, etc. \cite{Zha,Tawari13,wu2017locality}. 
The main challenge of the FER is to account for large appearance changes of human poses in unconstrained environment. A majority of the proposed methods were evaluated with constraint frontal FER, however, it remains a difficult task for developing robust algorithms to recognize multi-view facial expression in unconstrained environment with challenging factors such as pose variations, displacement of facial parts, non-linear texture warping, and self-occlusion ~\cite{Lopesa17,Dapogny16,Jampour2017,li2018deep}. To address this issue, we propose an end-to-end trainable pose-adaptive hierarchical attention network (PhaNet) that can jointly recognize the facial expressions and poses under multi-view and unconstrained environment with great efficiency and robustness. 
\begin{figure*}[t]
\centering
\centerline{\includegraphics[width=6in]{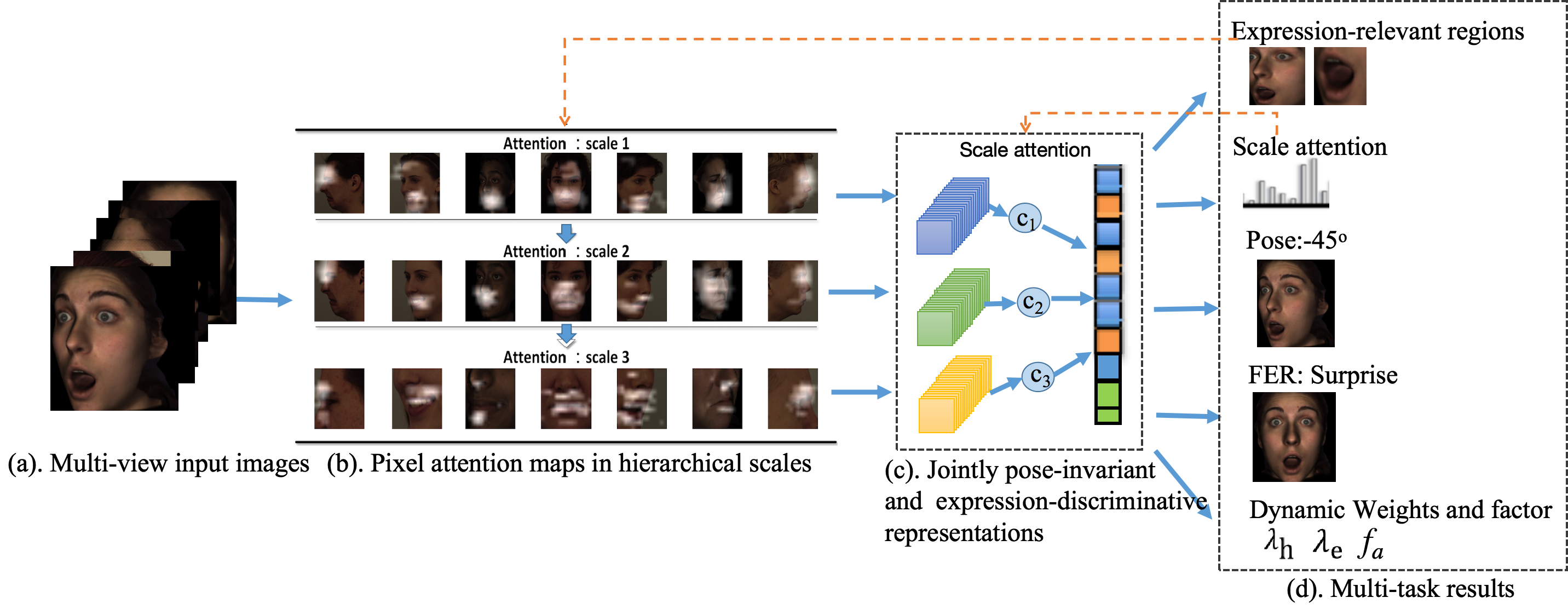}}
\caption{The overview of the PhaNet for Muti-view FER. It includes there parts, i.e., hierarchical pixel attention learning (HPAL), joint scale attention learning (SAL), and dynamically constraint multi-task learning (DCML). (a) Multi-view face images as the input, (b) the learned attention maps by HPAL in hierarchical scales, (c) the learned pose-invariant and expression-discriminative representations by joint SAL, (d) the recognized multi-task results by the DCML. Noted that the red dashed represents the feedback of the recurrent results.}\label{fig:motivation}
 \end{figure*}
 
It is not easy to perform multi-view FER in unconstrained environment. The main challenging factors for multi-view FER have two-folds. The one is the more discriminative representations between different poses than different expressions \cite{liu2018multi}. The another challenging comes from self-occlusion and facial texture warping \cite{ding2017pose}, which means there is loss of information for FER. 
To solve these challenges, the existing methods for multi-view FER usually include  three perspectives, i.e., pose-robust feature extraction, pose normalization, and pose-specific classification. Pose-robust feature extraction based methods reply on hand-crafted visual invariant features, such as scaled-invariant transform (SIFT) or local feature points \cite{ZhengW14, Jung15}, which have the limited power in the challenge of nonlinear facial texture warping. Pose normalization based methods align a posed face to a frontal face based on a transform or generative adversarial Network (GAN) before conducting FER \cite{yin2018multi}. Pose-specific classification based methods train multiple classifiers at specific poses, which need more well-designed classifiers trained on the large amount of training data and may only handle discrete pose variations\cite{Zha, Dapogny16}. These methods usually have two stages and rely on good performance of pose estimators.  
Therefore, how to alleviate the influence of pose estimators and promote FER under multi-view and unconstrained environment is still significant research challenge. 

To address the above limitations of current methods in multi-view FER, we propose an end-to-end trainable network with a hierarchical attention mechanism -- PhaNet for adaptively exploring regions based expression-discriminative and pose-invariant representations. According to human vision, when global features are deformed or occluded, human visual systems tend to choose hierarchical local features to discriminate \cite{li2019occlusion}. 
Inspired by this intuition, PhaNet is designed to contain three parts for multi-view FER: hierarchical pixel attention learning (HPAL), joint scale attention learning (SAL), and dynamically constraint multi-task learning (DCML). 
Specifically, the HPAL part detects the most relevant regions to the facial expression without any region annotation by two cascaded weakly-supervised pixel attention networks (PANs) in hierarchical scales. The found rough scales are useful for learning global representations, and are prone to deterioration due to pose variation. The finer scales are advantageous for the pose-invariant representation and easily lack of information. After the scales are obtained, the joint SAL part selects the most informative scales to learn the pose-invariant and expression-discriminative representations according to their contributions. Finally, the DCML part is used to optimize the whole network and achieve multi-task results in a joint classification and regression way. Fig. \ref{fig:motivation} illustrates the overview of the PhaNet for multi-view FER. The PhaNet is an end-to-end trainable network without any pre-trained models and facial region annotation for achieving the best promotion of both pose and FER tasks. 
 
PhaNet aims at improving accuracy of multi-view FER and achieving the pose-invariant and expression-discriminative representations based on relevant regions, and its contributions include the following:
\begin{enumerate}
\item  we propose an end-to-end trainable PhaNet with three components: HPAL, joint SAL and DCML, to jointly recognize facial expressions and poses in a mutually reinforced way. The evaluation on five typical and challenging multi-view facial expression datasets showed it advantages over the state-of-the-art methods.
\item  we propose hierarchical attention losses to optimize weakly-supervised HPAL for discovering expression-relevant regions in hierarchical scales. Visualized results show that HPAL is effective in perceiving the most relevant regions to the facial expression.
\item  we introduce a joint scale attention model to select the most informative regions for further learning the pose-invariant and expression-discriminative representations. The joint SAL is optimized by a dynamically constraint multi-task manner and is capable of learning a low weight for a pose-variant region and a high weight for an pose-invariant and expression-discriminative one. 
\end{enumerate}

The rest of this paper is organized as follows:  Section \ref{sec:work} gives related work. Section \ref{sec:PhaNet} presents our PhaNet approach for multi-view FER. Section \ref{sec:Exp} discusses the experimental results using publicly available datasets. Section  \ref{sec:Conc} concludes this paper.

\begin{figure*}
\begin{center}
\centerline{\includegraphics[width=7in]{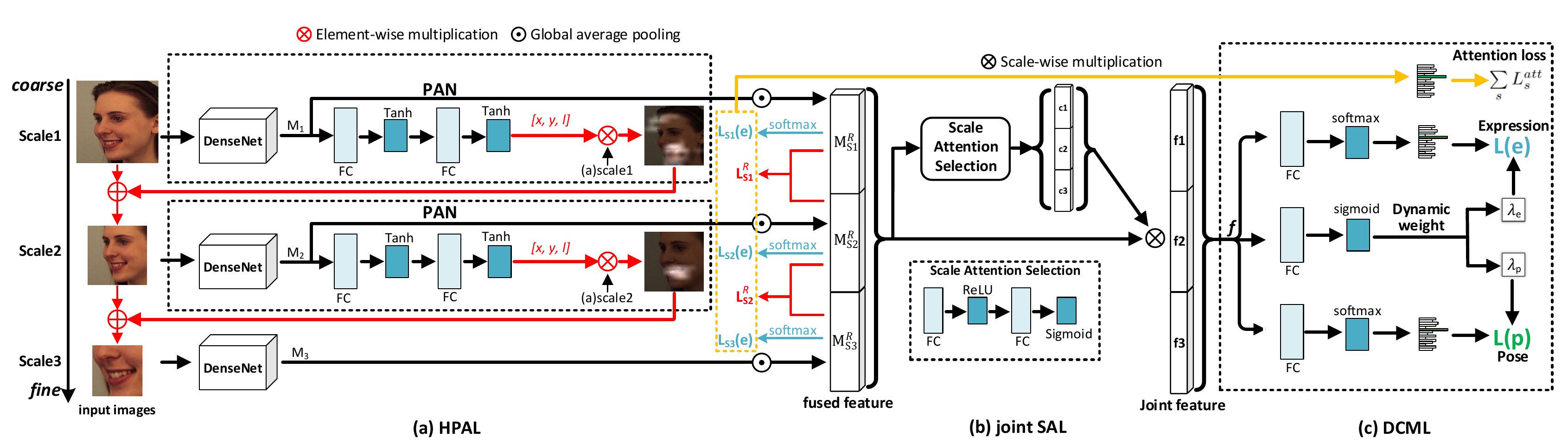}}
\end{center}
   \caption{The architecture of PhaNet with three components (i.e., (a) hierarchical pixel attention learning (HPAL), (b) joint scale attention learning (SAL), and (c) dynamically constraint multi-task learning (DCML)) for multi-view FER. Given an input image, first, a backbone CNN is used to compute its convolutional maps. Then taking the maps as input, the HPAL component with two pixel attention networks (PANs) generates the most relevant regions to the facial expression in hierarchical scales by minimizing the proposed hierarchical attention losses. After the regions are obtained, joint SAL selects the informative regions to further learn pose-invariant and expression-discriminative representations by minimizing a dynamically weighting multi-loss including the attention losses, pose loss and FER. Finally, fully-connected layers with the joint representations determine facial expressions, poses and the dynamic weights in a DCML manner. The whole network is optimized in an end-to-end way.} 
\label{fig:net}
\end{figure*}

\section{Related work} \label{sec:work}
In this section, we mainly discuss methods that are related
to multi-view FER,  attention models, and related network architectures.

 \textbf{Multi-view FER}
 Comparing to frontal FER, non-frontal FER
 is more challenging and more applicable in real scenarios.  Among existing works, pose-robust feature extraction, pose normalization and pose-specific multi-classifiers are the three most important categories of methods \cite{Lopesa17,Zha,ZhengW14,Jung15,Dapogny16,Rudovic10,yin2018multi}.~\cite{Dapogny16} proposed pair conditional random forests (PC-RF) to capture low-level expression transition patterns on the condition of head pose estimation for multi-view dynamic facial expression recognition. On the multi-view BU3D-EF dataset, the average accuracy reached 76.1\%. To reduce head pose influence, ~\cite{Jung15} trained a jointly CNNs with facial landmarks and color images, which achieves 72.5\%, and it contains three convolutional layers and two hidden layers. The higher accuracies are achieved with SIFT using the deep neural network (DNN)~\cite{ZhengW14,Zha}, which are 78.9\% and 80.1\% separately.~\cite{Lopesa17} proposed a combination of convolutional neural network (C-CNN) and special image pre-processing steps to recognize six expressions under head pose at 0$^\circ$ and achieved an averaged accuracy of 90.96\% on the BU3D-EF dataset.
  It's noted that head poses have large influence for FER in an unconstrained environment. ~\cite{yin2018multi} proposed a GAN based structured method for pose-invariant FER and simultaneous facial image synthesis, and achieved state-of-the-art performance due to the training set enlargement. The good results have been achieved by well-designed features based classifiers and large amount of training data, however, there is still a certain gap in multi-view unconstrained environment with limited amount of training data. How to address pose-adaptive FER with limited amount of data and multi-view unconstrained environment for improved performance is still an open problem.  
  
\textbf{Attention model} 
 Visual attention based networks have been proposed to localize significant regions for many computer vision tasks, including fine-grained recognition \cite{fu2017look, hu2018squeeze}, image caption \cite{xu2015show}, person re-identification \cite{zhao2017deeply}, and object detection \cite {ren2015faster}. Some methods are learned by clustering scheme from the internal hidden representations in CNN \cite{zheng2017learning}. Another methods focus on detecting local regions according to supervised bounding box annotation, e.g.,region proposal network (RPN) \cite{ren2015faster}. 
Moreover, Zheng et al. \cite{zheng2017learning} adopted channel grouping sub-network to cluster different convolutional feature maps into part groups according to peak responses of maps, which do not need part annotations. Xu et al. \cite{xu2016deepgender} proposed attention shift based on multiple blur levels to avoid occlusions for facial gender classification. SENet \cite{hu2018squeeze} proposed a novel architectural unit termed as Squeeze-and-Excitation (SE) block which adaptively recalibrates channel-wise feature responses by explicitly modeling interdependencies between channels. It can produce considerable performance improvement for image classification with minor additional 
computational cost.

Our model superficially resembles attention networks trained without any region annotation instead of a general RPN. The most relevant work to our model comes from \cite{fu2017look, hu2018squeeze}. 
The important differences are that the propose PhaNet can discover the most relevant regions to the facial expression in hierarchical scales and learn the pose-invariant and expression-discrimination representations from the scales by joint pixel and scale attention mechanism. In particular, our approach can adaptively assign a higher weight for a pose-invariant and informative facial region according to the pose loss and FER loss. 
     
 \textbf{Related network architectures}
 Our network architecture is inspired from multi-task network \cite{ranjan2016hyperface}, multi-scale network \cite{yin2018multi, liu2018multi}
 and densely connected convolutional network \cite{huang2017densely} to rapidly construct a low-resolution feature map that is amenable to multi-task classification. Ranjan \emph{et al.} \cite{saxena2016convolutional} proposed a deep CNN followed by a multi-task learning algorithm with fixed empirical weights. Huang \emph{et al.} \cite{huang2017densely} designed the same feature-concatenation approach, which allows features optimized for early classifiers in later layers of the network. Different from the related \cite{yin2018multi} and \cite{huang2017densely}, our model includes jointly attention networks with dense connection in a hierarchical way and a multi-task network with dynamically learned loss weights. It can suppress over-fitting and gradient disappearing for the best coupling. 

\section{Proposed approach}\label{sec:PhaNet}

In this section, we first give a brief overview of the proposed end-to-end PhaNet
for multi-view FER, then describe the learning process of each part in our approach.

 \subsection{PhaNet framework}
Our proposed framework PhaNet is shown in Fig. \ref{fig:net}, which consists of three parts: hierarchical pixel attention learning (HPAL), joint scale attention learning (SAL), and dynamically constraint multi-task learning (DCML).  
Given a facial image $X$ with the label $y=\{y^e, y^p\}$, where $y^e$ represents the label for the facial expression and $y^p$ for the pose, the goals of our learning problem are threefold: 
(1) HPAL discoveries the most relevant regions as a square $R_s(x, y, l) $ to the facial expression in  three hierarchical scales $s={S_1, S_2, S_3}$, where $ x, y, l $ denote the square's center coordinates and the half of the square's side length of the region. 
(2) Joint SAL selects the most informative scales to learn the pose-invariant and expression-discriminative representations $f$ according to their scale attention $c$.
(3) DCML predicts facial expressions $\hat{\mathbf{e}}$ and variant poses $\hat{\mathbf{p}}$ based on the joint representations. 

From the Fig. \ref{fig:net}, the PhaNet is optimized to convergence by alternatively learning the FER loss at each scale, an attention ranking loss across neighboring scales and a dynamically constraint multi-loss in the joint scale. Next we will introduce them in details.
  
  \subsection{HPAL for the most relevant region discovery} 
  
To discover the most relevant regions to the facial expression, the proposed hierarchical pixel attention learning (HPAL) part computes region attention in hierarchical scales. The architecture of HPAL consists of three backbone DenseNets \cite{huang2017densely} for feature concatenation and two weakly-supervised pixel attention networks (PANs) for hierarchical region localization (see Figure \ref{fig:net}). 

\subsubsection{Dense convolutional feature concatenation} 
To strengthen feature propagation and reduce the number of parameters, densely convolutional operation is applied to extract convolutional feature maps $\bf{M_s}$ by the dense connectivity as DenseNet at each scale. The densely deep feature receives the feature-maps of all preceding layers, 
  \begin{equation}
  {\bf{M_s}} = {{\mathop{\rm H}\nolimits} }([{m_s^0},{m_s^1},...{m_s^{l }}]),
   \label{eq:dense}
   \end{equation}
  where $m_s^0,..., m_s^l$ refer to the feature-maps produced in layers $0,...,l$ by using convolution operation at each scale and ${\rm{H(\bullet )}}$ denotes a concatenation operation. 
  
\subsubsection{Hierarchical region localization} 
 Different from RPN in object detection which uses strong supervision of ground truth boxes, two weakly-supervised PANs discover the most relevant regions to the facial expression by minimizing the expression loss and an attention ranking loss. We further model each PAN as a multi-task learning with two outputs, i.e., the predicted expression probability $p(e|s)$ at the current scale and a set of box coordinates $ R_s(x, y, l)$ of an attended region for the next finer scale. To discover the attended region as a square with three coordinate parameters, the representation is given by,
    \begin{equation}
      [{x},{y},{l}, p(e|s)] = \Upsilon ({\bf{M}_s}),
     \label{eq:attented feature}
     \end{equation}
 where  $x$ and $y$ denote the square's center coordinates, respectively, and $l$ denotes the half of the square's side length. The specific form of $\Upsilon (\bullet )$ can be represented by two-stacked fully-connected layers at the $S^{th}$ scale. 

At the first scale $S_1$, the square is selected by searching regions in the original image, with the highest response value in the last convolutional layer of DenseNet. For the finer region discovery at the next scale $S_2$, PAN uses the expression probability $p(e|s)$ of the current scale to guide the next smaller square localization, which preferably makes the optimized localization more relevant to the facial expression. Similarly, two weakly-supervised PANs respectively approximate the most relevant regions at the second scale $S_2$ and the third scale $S_3$ by minimizing the proposed attention ranking loss and expression loss as:
  \begin{equation}
      {L_{s}^{att}} =   {L_s^{R}} (p(e,{y^e} | s),p(e,{y^e} | s-1))+ L_{s}(M_s, y^e)
      \label{eq:attend_loss}
      \end{equation}
 where $p(e,{y^e} | s)$ denotes the prediction probability on the correct expression categories $y^e$. $L_{s}(M_s, y^e)$ is the expression loss at the $S^{th}$ scale, which guides the region shifting (see the abbreviation $L_s(e)$ in Figure~\ref{fig:net}(b)). The proposed attention ranking loss $L_s^{R}$ is given by:
   \begin{equation}
   L_s^{R} = \max\{ 0, p(e,{y^e} | s-1) - p(e,{y^e} | s) + mar\},
   \end{equation}
 where $mar$ is a margin as 0.05 which makes the loss less sensitive to noises.  It can enable the PAN to take the prediction from coarse scales as references, and gradually approach the most relevant region by enforcing the finer-scale network to generate more confident predictions.
 
 In order to optimize coordinates, each PAN calculates the derivative on $x, y, l$ by the chain rule in backward-propagation and shows the effects to region detection. 
For the minimization of the attention loss, if the $\frac{{\partial L_s^{R}}}{{\partial {x}}}  \prec 0$, $x$ will increase, otherwise $x$ will decrease. The iteration will be continue until the ${L_{s}^R}$ is the lowest. The same operations have been happened in the derivatives on $y$ and $l$. 

 \subsubsection{Hierarchical region cropping} 
 Once the location of an attended region is approximated, we crop and zoom in the attended region to finer scale with higher resolution to extract more discriminative features. We approximate the cropping operation by proposing a variant of two-dimension boxcar function as an attention mask. 
 The mask can select the highest response regions in forward-propagation, and is to be optimized in backward-propagation. 
 The cropping operation can be performed by an element-wise multiplication between the original image at coarser scales and the attention mask, which can be computed as:
     \begin{equation}
     \begin{array}{l}
     {\bf{R}}_{s }^{att} = {\bf{R}}_{s-1}^{att} \otimes {\bf{T_a}}({\bf{M}_s}({x},{y},{l})),{\kern 1pt} {\kern 1pt} {\kern 1pt} {\kern 1pt} {\kern 1pt} {\kern 1pt} {\kern 1pt} {\kern 1pt} {\kern 1pt} {\kern 1pt} {\kern 1pt} {\kern 1pt} \\
     \end{array}
       \end{equation}
 where $\otimes$ represents element-wise multiplication, ${\bf{R}}_{s}^{att}$ denotes the cropped attended regions at the current scale, and ${\bf{R}}_{s-1}^{att}$ is the coarser region in the prior scale. $\bf{T_a}( \bullet )$ is the Tanh function to make the value range of $M_s$ from -1 to 1 and acts as an attention mask. Then, we further zoom the region by a bilinear interpolation way.
    
 \subsection{Joint SAL for the pose-invariant and expression-discriminative representations} 
\begin{figure}[H]
\centering
\centerline{\includegraphics[width=2in]{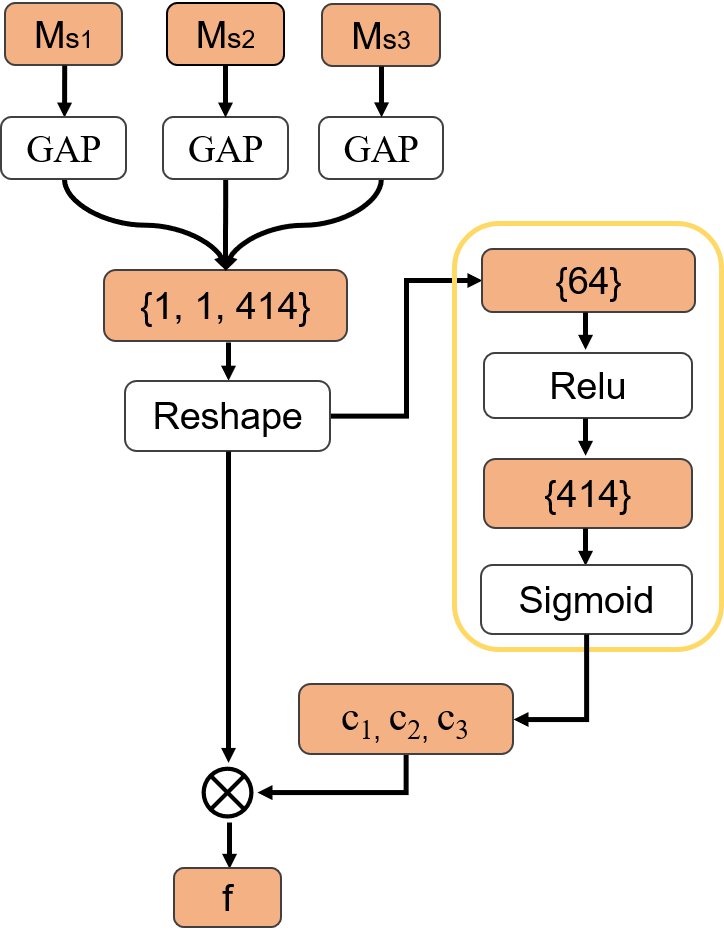}}
\caption{The structure of the joint scale attention learning network. It re-weights the region-specific representations according to posture invariance and expression discrimination. The GAP is an abbreviation of global average pooling. }\label{fig:sal}
 \end{figure}
 
 The joint SAL architecture is shown in Fig.~\ref{fig:sal}. Rather than extract features from the whole face as the traditional CNNs do, joint multi-scale features are learnt from the hierarchical regions by a scale-wise attention mechanism. The feature map of each scale can be regarded as a region-specific expressive representation. The  joint scale attention learning (SAL) network re-weights the region-specific representations according to posture invariance and expression discrimination by a scale attention mechanism. 

\begin{figure*}[!h]
\centering
\centerline{\includegraphics[width=5.5in]{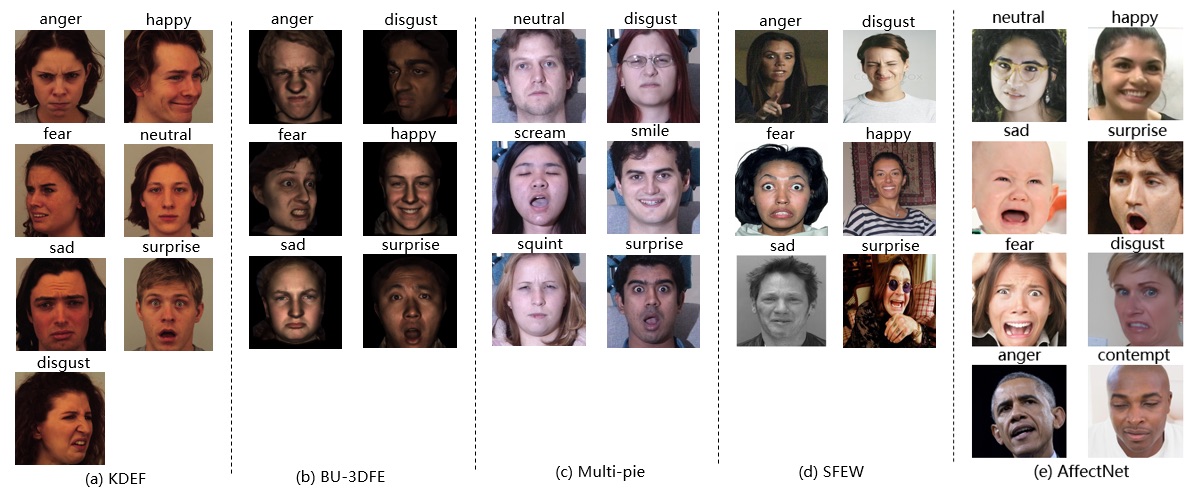}}
\caption{Examples of facial expression images from the five datasets. From left to right:
(a) KDEF, (b) BU-3DFE, (c) Multi-pie, (d) SFEW, (e) AffectNet.}\label{fig:dataset}
 \end{figure*}

The scale attention map can be calculated by a small 4-layers squeeze-and-excitation component. We first perform a squeeze operation via an global average pooling layer (zero parameters) for aggregation region-specific features into a preliminary fusion features. The preliminary fused features $\bf{M_F}$  can be represented as follows:
  \begin{equation}
    \begin{array}{l}
  {\bf{M_F}} = {{\mathop{\rm H}}} ( [\bf{M}_{S_1}^R,  \bf{M}_{S_2}^R, \bf{M}_{S_3}^R)],\\
   \bf{{M}_{S}^R} = g(\bf{{M}_{S}}),   
     \end{array}
   \label{eq:dense}
   \end{equation}
 where $g(\bullet)$ is the global average pooling operation. $\bf{M_{S_1}^R}$, $\bf{M_{S_2}^R}$, and $\bf{M_{S_3}^R}$ respectively denote the region-specific representations extracted from the hierarchical scales. ${\rm{H(\bullet )}}$ is the concatenation operation.

In the multi-excitation step, two fully connected layers are used to approximate the scale attention model with three coefficients as follows,
    \begin{equation}
  \begin{array}{l}
[{c_{1}},{c_{2}},{c_{3}}] = \sigma ( {\mathop{\rm ReLU}\nolimits} ({\bf{W}}_2^l \times {\mathop{\rm ReLU}\nolimits} ({\bf{W}}_1^l \times {{\bf{M}}_{\bf{F}}}))),\\
\end{array}
     \label{eq:channel feature}
     \end{equation}
 where the specific form of $\sigma$ refer to the Sigmoid function. 
 Three outputs ${c_{1}},{c_{2}},{c_{3}}$ respectively measure the three scales' impact on pose variation and encode a non-mutually-exclusive relationship among scales.  
 ${\bf{W}}_1^l \in {{\bf{R}}^{^{r \times 1}}}$ (r parameters) and ${\bf{W}}_2^l \in {{\bf{R}}^{^{\frac{r}{m} \times 1}}}$ ($\frac{r}{m}$ parameters) denote the parameter vector of 2 fully-connected layers in order respectively. $r $ represents the bottleneck reduction rate.
This step tends to learn a low weighting coefficient for a pose-variant scale and a high weighting coefficient for a pose-invariant and expression-discriminative one. 
 
Once the coefficients are learnt, we re-weight the preliminary fused representation $M_F$, and achieve the final jointly  multi-scale representation vector $\bf{f} \in {\bf{R}^{n \times 1}}$. A scale-wise multiplication between each region-specific representation $\bf{M}_s^R$ and the scale coefficient is performed as, 
  \begin{equation}
  {\bf{f}} = {\rm{H}}([{c_{1}} \times {\bf{M}}_{{\bf{S_1}}}^{\bf{R}},{c_{2}} \times {\bf{M}}_{{\bf{S_2}}}^{\bf{R}},  {c_{3}} \times {\bf{M}}_{{\bf{S_3}}}^{\bf{R}}]).
     \label{eq:scale}  
     \end{equation}
The equation shows that the jointly multi-scale representations are weighted sum of the region-specific features in hierarchical scales. Furthermore, the joint SAL models the pose-invariant and expression-discriminative dependencies between feature maps by optimizing a DCML manner.

 \subsection{DCML for classification and regression} 

Different from empirical fixed weights in traditional multi-task learning, dynamically constraint multi-task learning (DCML) introduces a dynamically weighting multi-loss function according to each task's contribution, for jointly classification and regression.
It consists of one full-connected layer, two Softmax layers for facial expression and pose classification respectively and one Sigmoid layer for dynamic weight regression. Moreover, a weighting constraint factor is designed to avoid gradient disappearance in training procedure.  

Specifically, the dynamically weighting multi-loss function includes a weighting expression recognition loss $L(e)$, a weighting pose estimation loss ${L(p)} $, the hierarchical attention loss $L_s^{att}$ as Eq. \ref{eq:attend_loss} and a weighting constraint factor. Our model's aim is to minimize the multi-loss of all tasks together, which is defined as:
   \begin{equation}
\begin{array}{l}
{L_{Multi}} = {\lambda _{\rm{e}}}L({\bf{f}},{{\bf{y}}^{\bf{e}}}) + {\lambda _{\rm{p}}}L({\bf{f}},{{\bf{y}}^{\bf{p}}}) +\sum\limits_{s } {L_s^{att}} + {e^{ - 12{\lambda _{\rm{p}}}(1 - {\lambda _{\rm{p}}})}},     
         \\ 
s.t.{\kern 1pt} {\kern 1pt} {\kern 1pt} {\kern 1pt} {\kern 1pt} {\kern 1pt} {\kern 1pt} {\kern 1pt} {\kern 1pt} {\kern 1pt} {\kern 1pt} {\kern 1pt} {\lambda _{{\rm{p}}}} + {\lambda _{{\rm{e}}}} = 1,{\kern 1pt} {\kern 1pt} {\kern 1pt} {\kern 1pt} {\kern 1pt} {\kern 1pt} {\kern 1pt} {\lambda_{\rm{e}}}\ge {\lambda _{{\rm{p}}}}\ge {\rm{0}},  {\kern 1pt} {\kern 1pt} {\kern 1pt}  s= \{S_2, S_3\}
\end{array}
  \end{equation}
 where ${\lambda_{\rm{e}}}, {\lambda _{\rm{p}}}$ are dynamic weights that control and assign the contributions of expression and pose tasks. The designed weighting constraint factor ${{e^{ - 12{\lambda _{\rm{p}}}(1 - {\lambda _{\rm{p}}})}}}$ can help to suppress gradient disappearance during the fused training due to the loss discrepancy between the two tasks. We set the sum of the two tasks' weights  to 1, and learn the dynamic weights as follows:
     \begin{equation}
     [{\lambda _{\rm{e}}},{\lambda _{\rm{p}}}] = \sigma({{\bf{W}}^\lambda }{{\bf{f}}} + {{\bf{b}}^\lambda }),
       \end{equation}
where $\sigma$ refers to Sigmoid function. $\bf{W^\lambda }$ and $\bf{b^\lambda }$ denote the network's weight matrix and bias vector in this fully connected layer. Since FER task is the main task, we impose the constrained condition ${\lambda _{\rm{e}}} \ge {\lambda _{\rm{p}}}$. Simultaneously, regions based the joint multi-scale representations can be also updated by minimizing the dynamically weighting multi-loss.


\section{Experimental Results} \label{sec:Exp}
To evaluate our approach, five face expression datasets were used: KDEF dataset \cite{lundqvist1998karolinska},  BU-3DFE dataset \cite{Yin06}, Multi-pie dataset \cite{gross2010multi-pie}, AffectNet \cite{kollias2018affwild2} and SFEW in-the-wild dataset \cite{dhall2015video}. The examples from the five datasets are shown in Fig.~\ref{fig:dataset}

 \subsection{Datasets}
\textbf{BU-3DFE:} The  BU-3DFE contains 100 people of different ethnicities, including 56 females and 44 males. Seven facial expressions (Anger(AN), Disgust (DI), Fear (FE), Happiness (HA), Sadness (SA), Surprise (SU) and Neutral (NE)) are elicited by various manners and nine pan angles $-90^\circ$, $-ˆ'60^\circ$,$-45^\circ$, $-30^\circ$, $0^\circ$, $30^\circ$, $45^\circ$, $60^\circ$and $90^\circ$, and each of them includes 4 levels of intensities. 

\textbf{Multi-pie:} The Multi-pie is for evaluating facial expression recognition under pose and illumination variations in
the controlled setting. We use images of 270 subjects depicting acted facial expressions of Neutral (NE), Disgust (DI), Surprise (SU), Smile (SM), Scream(SC), and Squint (SQ), captured at five pan angles $-30^\circ$, $-15^\circ$, $0^\circ$, $15^\circ$ and $30^\circ$, resulted in 1531 images per pose.

\textbf{KDEF:} The KDEF contains 35 females and 35 males displaying 7  emotional expressions (Anger(AN), Disgust (DI), Fear (FE), Happiness (HA), Sadness(SA), Surprise (SU) and Neutral (NE)). Each expression is viewed from 5 different pan angles $-90^\circ$, $-ˆ'45^\circ$, $0^\circ$, $45^\circ$ and $90^\circ$.

\textbf{AffectNet:} The AffectNet contains around 420,000 annotated images and each image is labeled by only one human coder. It includes 5,500 labeled images in 10 categories as the validation set. We use images of 8 expression categories (Anger(AN), Disgust (DI), Fear (FE), Happiness (HA), Sadness(SA), Surprise (SU), and Contempt (CO)), and have labelled the pan angle of each image (from $-45^\circ$ to $45^\circ$) by using the Openface method ~\footnote {http://cmusatyalab.github.io/openface/} \cite{amos2016openface}.

\textbf{SFEW:} The SFEW dataset is a subset of  EmotiW2015 in-the-wild emotion dataset, which contains static images based spontaneous facial expression collected in real-world conditions. In our experiments, we only use SFEW in cross evaluation.

\subsection{Implementation details}

We construct the network according to Fig. \ref{fig:net}. We first resize input images as $224 \times 224$. The training and validation data sets include 16812 images with 75 persons from BU-3DFE dataset, 3919 images with 56 persons from KDEF dataset, and 7370 images from Multi-pie dataset, 37,000 images from AffectNet. A 5-fold cross-validation was conducted for parameter adjustment.  
For testing, we used  other 979 images from KDEF dataset,  3852 images from the BU-3DFE dataset, 1840 images from Multi-pie dataset, 5000 images from AffectNet and 755 images from SFEW dataset. We guarantee that the persons in training and testing procedures are independent subjects.

We used the Tensorflow framework \cite{abadi2016tensorflow} for implementing CNN. The important training parameters in the experiments include initial learning rate (0.01),  learning rate delay(0.1), mini-batch size(32), convolution kernel size in the $1^{st}$ layer (7${\times}$7), convolution kernel sizes in three dense blocks (3${\times}$3 and 1${\times}$1), and neural nodes in full connected layers (128, 64, 512). All DenseNet convolutional modules are trained for Batch Normalization (BN) with $k=12$ and $depth=40$,  The experiments were conducted on a PC with Intel (R) Core(TM) i7-6700 CPU at 4.00GHz and 32GB memory, and NVIDA GeForce GTX 1080. In order to evaluate our model effectively, we will release our source code to Github in future~\footnote {https://github.com}.


\subsection{Experiments with BU-3DFE Dataset}

\begin{figure}[!ht]
\centering
\centerline{\includegraphics[width=2.5in]{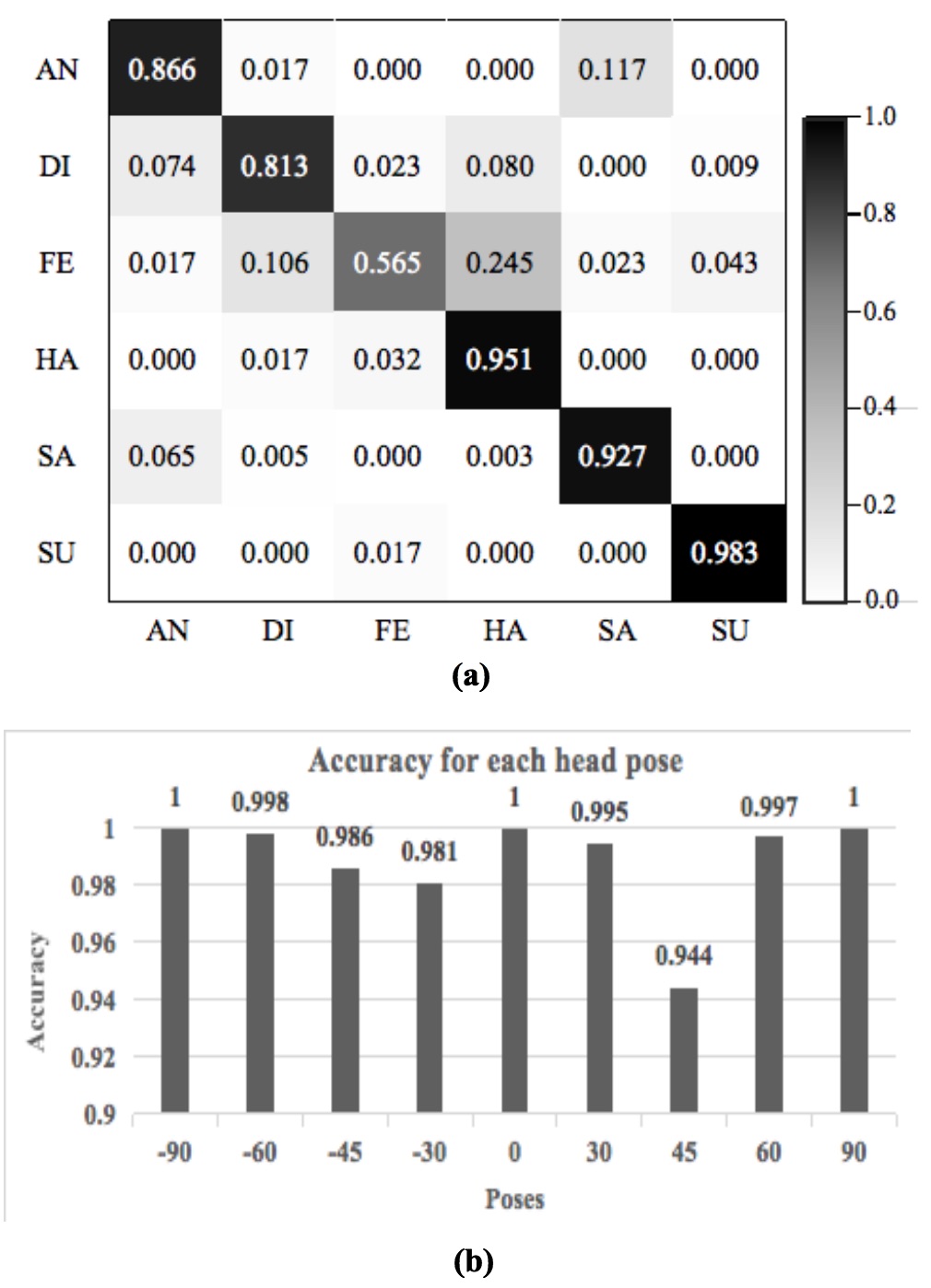}}
\caption{Overall performance on BU-3DFE dataset. (a)The confusion matrix for FER, (b)the accuracy for each head pose
}\label{fig:bu3d}
 \end{figure}

 In Fig.~\ref{fig:bu3d}(a) and Fig.~\ref{fig:bu3d}(b), we show the confusion matrix for FER and the accuracy for each head pose by using our method. 
 Among the six expressions, there are three expressions (HA, SU, and SA) with higher accuracies over 90\%. The highest accuracy is 98.3\% of SU while the lowest accuracy is 56.5\% for FE, which has the least amount of facial movement and is difficult to distinguish with other expressions. The average accuracy of expression recognition is 84.92\% under overall head poses, and the average accuracy of head poses is 99.49\%.

 \begin{table}[!ht]
\centering
\caption{Comparison of accuracies (\%) using different methods for multi-view FER on BU-3DFE. The highest results are highlighted in bold.}
\resizebox{0.48\textwidth}{!}{%
\begin{tabular}{c|cc|cc} 
\hline
Methods                                                       & Features  & Pan  &  Acc. of pose & Acc. of FER \\\hline\hline
\textbf{PhaNet  }                                      &{ Pixel and scale attention } & { (-90$^\circ$,90$^\circ$)}& \bf{99.49} &\textbf{84.92} \\
DenseNet~\cite{huang2017densely}                  & Dense features & (-90$^\circ$,90$^\circ$) & 94.45 &80.39 \\ 
PC-RF ~\cite{Dapogny16}                              & Heterogeneity &   (-90$^\circ$,90$^\circ$) & 87.15 &  76.1 \\
JFDNN ~\cite{Jung15}                              &Image and landmarks &  (-90$^\circ$,90$^\circ$) & - & 72.5   \\
GSRRR~\cite{ZhengW14}                      &Sparse SIFT &  (-90$^\circ$, 90$^\circ$) & 87.36 &78.9  \\
DNN-D~\cite{Zha}                                        & SIFT   &  (-90$^\circ$, 90$^\circ$) & 92.26  & 80.1  \\
SVM~\cite{Moore11}                                  & LBP and LGBP &(-90$^\circ$,90$^\circ$) & - &71.1  \\ 
\hline
\textbf{PhaNet  }                                      &{Pixel and scale attention} & { (-45$^\circ$, 45$^\circ$)} & \bf{99.43} & \textbf{84.74} \\
GAN ~\cite{zhang2018joint}      & Convolutional features & (-45$^\circ$, 45$^\circ$) & 95.38 & 81.2  \\ 
LLRS\cite{jampour2015multi}                           & Sparse features & (-45$^\circ$, 45$^\circ$) &- & 78.64    \\ 
SSE\cite{tariq2014supervised}                                &Supervised super-vector encoding & (-45$^\circ$, 45$^\circ$) &-& 76.60  \\ 
MMGL\cite{tariq2013maximum}                                 & Soft Vector Quantization & (-45$^\circ$, 45$^\circ$) &-& 76.34 \\
\hline
\end{tabular}}
\label {tab:comp_3D}
\end{table}

\begin{table*}[ht!]
	\centering
	\caption{Average accuracy  (\%) of  pose-aligned FER under 9 pans on the BU-3DFE dataset. The highest accuracy for each pose  is highlighted in bold.}
	\begin{tabular}{c|ccccccccc|c}
		\hline
		Methods & \multicolumn{9}{c|}{ Poses} & \multicolumn{1}{c}{Average}\\ \cline{2-10}
		                                & $-90^\circ$  &  $-60^\circ$  & $-45^\circ$ & $-30^\circ$       &  $0^\circ$  & $30^\circ$   &  $45^\circ$  & $60^\circ$ & $90^\circ$   &  \\\hline\hline
	   \textbf{PhaNet}                                   & \bf{84.0}  & \bf{86.3}	& \bf{84.8} & \bf{83.9 }     & \bf{85.4}	& \bf{84.4}  & \bf{85.6} & \bf{86.4} & \bf{83.5} &        \textbf{84.92}  \\
	    DenseNet~\cite{huang2017densely}                & 80.1& 81.7 & 81.8 & 80.0   & 79.2 & 81.1& 82.1 & 82.1 & 80.0 & 80.39 \\
		GSRRR~\cite{ZhengW14}                           & 77.3& 78.4. & 80.4 &79.5    & 78.9    & 80.1& 80.1& 78.4 &77.0 &  78.9 \\
    	DNN-D~\cite{Zha}                                    & 79.7& 80.5 & 81.2 &79.9   & 79.7   & 80.7& 81.0& 80.5 &79.51& 80.1 \\
		\hline
	\end{tabular}
	\label{tab:head_ffr}
\end{table*}

The average expression and pose accuracies of our PhaNet is compared with the state of the arts, including  GAN~\cite{zhang2018joint}, DenseNet~\cite{huang2017densely}, Pair conditional random forests (PC-RF)~\cite{Dapogny16}, Joint fine-tuning in deep neural networks (JFDNN)~\cite{Jung15},  Group sparse reduced-rank regression (GSRRR)~\cite{ZhengW14}, Deep neural network-driven SIFT (DNN-D)~\cite{Zha}, and SVM~\cite{Moore11} in Table~\ref{tab:comp_3D}. Specifically, the methods conduct the multi-view FER on a relatively set of discrete poses containing 5 pan angles \cite{Dapogny16,Jung15,Moore11} or 9 pan angles \cite{huang2017densely, ZhengW14, Zha} or 35 pan angels \cite{zhang2018joint, jampour2015multi, tariq2014supervised, tariq2013maximum}. 
Dapogny et. al.~\cite{Dapogny16} proposed PC-RF to capture low-level expression transition patterns on the condition of head pose estimation for multi-view FER. The average accuracy reached 76.1\%. JFDNN achieves 72.5\% on FER by using joint two fine-tuning networks. 
The higher accuracies of FER are achieved with GAN based structure~\cite{zhang2018joint} and DenseNet~\cite{huang2017densely}, which are 81.2\% and 80.39\%, respectively.  Our method achieves the highest accuracies of 84.92\% and 99.49\% on pose and FER,  which are competitive to the methods above. 


Table~\ref{tab:head_ffr} lists average FER accuracies under different head poses of the PhaNet, DenseNet~\cite{zhang2018joint}, GSRRR~\cite{ZhengW14} and DNN-D~\cite{Zha}.  The accuracies of our method are significantly greater than that of the other three methods in terms of different views.  
The highest accuracy of our method is achieved under the head pose 60$^\circ$, which is  86.4\%. And the lowest accuracies appear under the head pose 90$^\circ$ due to face deformation.

\subsection{Experiments with Multi-pie dataset}
\begin{figure}[!htb]
\centering
\centerline{\includegraphics[width=2.5in]{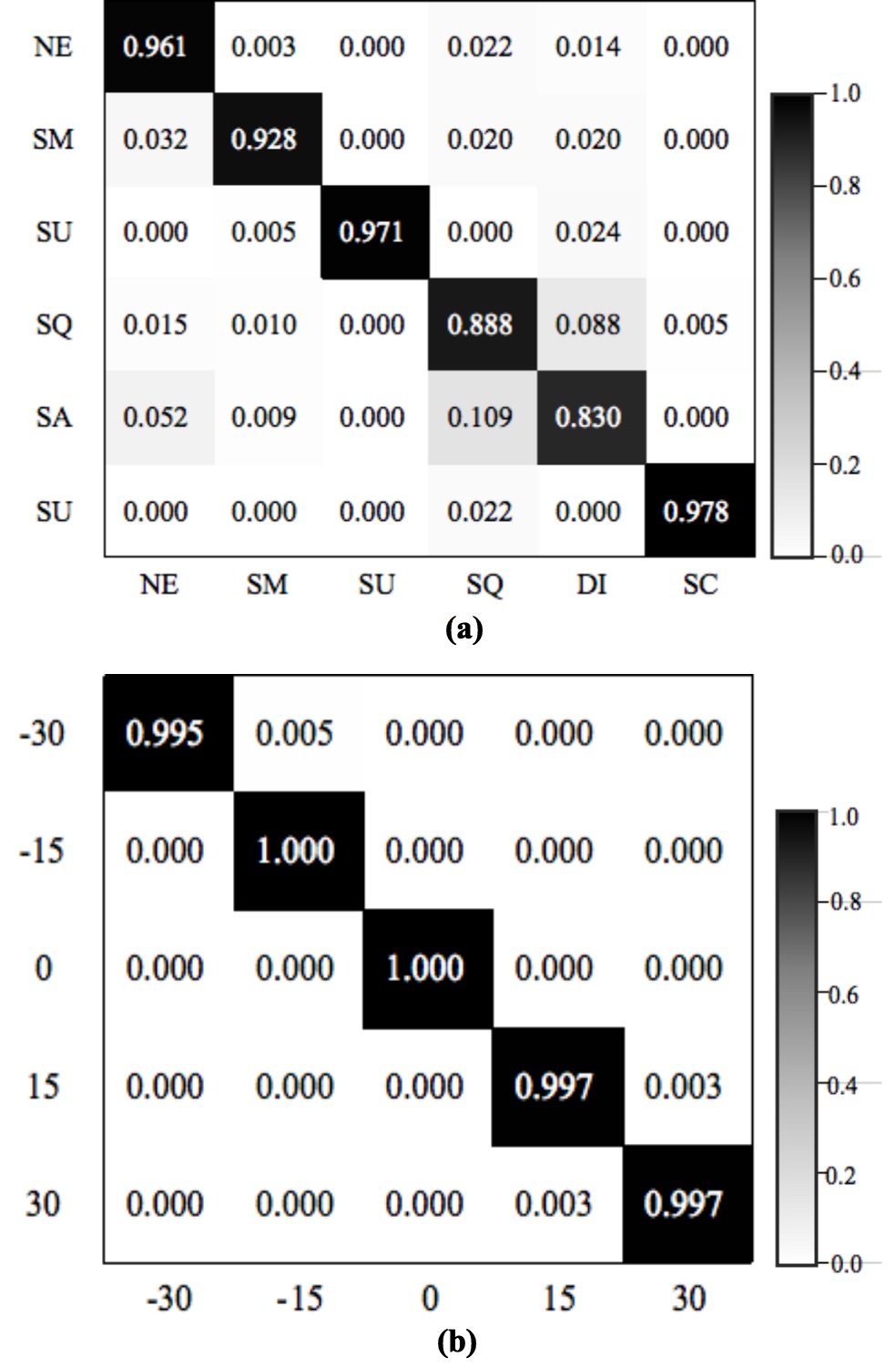}}
\caption{Overall performance on Multi-pie dataset. (a)The confusion matrix for FER,
(b) the confusion matrix for head pose estimation.}\label{fig:Multipie}
 \end{figure}

 In Fig.~\ref{fig:Multipie}(a) and Fig.~\ref{fig:Multipie}(b), we show the confusion matrixes for multi-view FER and head pose estimation by using our method. The average accuracy of expression recognition is 93.53\% under overall poses, which prove that our method can reduce the influence of pose variances. The average accuracy of pose is 99.78\%.

 \begin{table}[ht!]
 	\centering
 	\caption{Comparison of state-of-the-art methods on the Multi-Pie
 	dataset. The highest accuracy for each pose is highlighted in bold.}
 	\resizebox{0.49\textwidth}{!}{%
 	\begin{tabular}{c|ccccc|c}
 		\hline
 		Methods & \multicolumn{5}{c|}{ Poses} & \multicolumn{1}{c}{Average}\\ \cline{2-6}
 			                                                              
 		                                           &  $-30^\circ$  & $-15^\circ$   &  $0^\circ$  & $15^\circ$ & $30^\circ$  &    \\\hline\hline \cline{2-6}
 	   \textbf{PhaNet}                                             & \bf{93.59}    &	93.38&	\bf{93.75} &	\bf{93.59}	& \bf{93.38}  & \bf{93.53} \\
 	    GAN~\cite{zhang2018joint}    & 90.97 & 94.72 & 89.11 & 93.09 & 91.30 & 91.80\\
 	    DenseNet~\cite{huang2017densely}                & 90.00~ & 91.67~& 90.56~ & 91.67~ & 90.83  & 91.06\\
 	KNN~\cite{Stefanos2015}                                                                 & 80.88 & 81.74 & 68.36 & 75.03 & 74.78 & 76.15  \\
     	LDA~\cite{Stefanos2015}                                                            & 92.52  & 94.37 & 77.21 & 87.07 & 87.47 & 87.72\\
     	LPP~\cite{Stefanos2015}                                                               & 92.42 & 94.56 & 77.33 & 87.06 & 87.68 & 87.81 \\
     	D-GPLVM~\cite{Stefanos2015}                                                       & 91.65  & 93.51  & 78.70  & 85.96  & 86.04  & 87.17\\
     	GPLRF~\cite{Stefanos2015}                                                         & 91.65  & 93.77  & 77.59  & 85.66  & 86.01  & 86.93\\
     	GMLDA~\cite{Stefanos2015}                                                        & 90.47  & 94.18  & 76.60  & 86.64  & 85.72  & 86.72\\
     	GMLPP~\cite{Stefanos2015}                                                         & 91.86  & 94.13  & 78.16  & 87.22  & 87.36  & 87.74\\
     	MvDA~\cite{Stefanos2015}                                                            & 92.49  & 94.22  & 77.51  & 87.10  & 87.89  & 87.84\\
     	DS-GPLVM~\cite{Stefanos2015}                                                  & 93.55 & \bf{96.96}  & 82.42  & 89.97  & 90.11  & 90.60\\
 		\hline
 	\end{tabular}}
 	\label{tab:Multi-pie}
 \end{table}

We then evaluate our method by comparing its performance
with the current state-of-the-art methods reported
in ~\cite{Stefanos2015} including kNN, LDA, LPP, D-GPLVM, GPLRF,
GMLDA, GMLPP, MvDA, and DS-GPLVM, GAN based structure~\cite{zhang2018joint} and DenseNet~\cite{huang2017densely}. The detailed
results across all views are summarized in Table ~\ref{tab:Multi-pie}. The mean multi-view FER accuracy is reported in the last column. The results clearly show that our method outperforms all existing methods with a 17.16\% to 1.73\% improvement in terms of FER accuracy. 
There are mainly two reasons. The one is the more pose-invariant and expression-discriminative representations learnt by our model, and the other reason maybe deformation training samples generated by the GAN based structure. Note that other models cannot achieve
good performances in the front and various views. However, our model can significantly improve the performance attained by the images with arbitrary poses and expressions.

\begin{table}[ht!]
	\centering
	\caption{Results on the KDEF dataset in terms of the recognition
	rates (\%). The  top row  indicates different views, and the leftmost column indicates different
	facial expressions.}
	\resizebox{0.49\textwidth}{!}{%
	\begin{tabular}{c|ccccc|c}
		\hline
		Exp./pose                             &  $-90^\circ$  & $-45^\circ$   &  $0^\circ$  & $45^\circ$ & $90^\circ$  & Average    \\\hline\hline                    
	  Anger (AN)                                   & 82.1 & 	78.6 &	85.7&	67.9 &	82.1 &	79.3  \\
	  Disgust (DI)                             &   85.7 &	89.3 & 	96.4 &	89.3 & 82.1	& 88.6 \\
	  Fear (FE)                                  &    82.1 &	82.1 &	71.4 &	67.9 &	85.7 &	77.8 \\
	  Happiness (Ha)                                   & 100.0 &	100.0 &	100.0 &	100.0 &	100.0 & {100.0} \\
	  Sadness (SA)                                     & 96.3	 & 88.9	& 85.2	& 81.5	& 88.9	& 88.2 \\
	  Surprise (SU)                             & 92.6 &	92.6 &	88.9 &	85.2 &	92.3.6 &	90.3 \\
	  Neutral (NE)                                     & 92.6 &	100.0 &	100.0 &	92.6 &	92.6 &	95.6 \\	\hline
	  Average                                     & {90.2} &	{90.2} &	89.7 &	83.5 &	89.1 &	{88.5} \\                                          
		\hline
	\end{tabular}}
	\label{tab:KDEF}
\end{table}

\begin{table}[ht!]
\centering
\caption{Comparison of accuracies (\%) using different methods on KDEF with 5 pan pose angels. The highest accuracies are highlighted in bold.}
\resizebox{0.49\textwidth}{!}{%
\begin{tabular}{c|c|cc} 
\hline
Methods                                                       & Features  & Acc. on Poses   &  Acc. on FER  \\\hline\hline
\textbf{PhaNet  }                                      &{Pixel and scale attention} & \bf{100}& \bf{88.5} \\
DenseNet~\cite{huang2017densely}          & Facial image & 99.23 & 85.10\\ 
TLCNN~\cite{zhou2017action}                           & Action Unit Selective Feature & 97.55 & 86.43  \\
SURF boosting~\cite{rao2015multi}                  & SURF   & - & 74.05  \\
SVM~\cite{Moore11}                                   & LBP and LGBP  & 86.67 & 70.5   \\ 

\hline
\end{tabular}}
\label {tab:comp_KDEF}
\end{table}

\begin{table*}[!htb]
\centering
\caption{Comparison result ( \%) of each individual attention component in PhaNet on the four datasets. No attention ($N_{S_1}$), hierarchical pixel attention ($H_{S_2}, H_{S_3}$), joint scale attention ($J_{SA}$).}
\begin{tabular}{c|c|c|c|c}
\hline
Different attention & Acc. on BU3D & Acc. on Multi-pie & Acc. on KDFE & Acc. on AffectNet\\
\hline\hline
$N_{S_1}$ (No attention ) & 77.79  & 91.45& 85.31 & 49.90\\
$H_{S_2}$ & 83.10 &92.05 & 87.71&49.30 \\
$H_{S_3}$ & 80.57 & 85.25 & 80.31 & 44.31 \\ 
$N_{S_1}+H_{S_2}$ & 83.20 & 92.49 & 88.23 & 49.99\\
$N_{S_1}+H_{S_3}$ & 84.24 &93.42 & 87.29 &50.12\\
$H_{S_2}+H_{S_3}$ & 81.57 &87.91 & 84.23 & 49.68 \\
$N_{S_1}+H_{S_2} +H_{S_3} $ & { 84.74}& {93.31}& {88.35} & {48.56} \\
\hline
$N_{S_1}+H_{S_2 }+H_{S_3} + J_{SA}$ & \bf{ 84.92}& \bf{93.53}& \bf{88.5} & \bf{54.82}\\
\hline
 \end{tabular}
\label {tab:paths}
\end{table*}

\subsection{Experiments with KDEF dataset}

The results of KDEF dataset are shown in Table~\ref{tab:KDEF}. The bottom row
represents the average FER rates for different
views (a total of 5 views). The rightmost column represents
the average recognition rates for 7 different facial expressions, and
the bottom-right corner cell represents the average overall
recognition rate with multi-views. The results show that our method
achieves the average FER accuracy of 88.5\%. Furthermore,
among the 7 expressions, HA is easier to be recognized with accuracy 100\%. The two lowest recognized expressions are FE and AN. These two expressions have the least amount of facial movement to find the relevant regions and thus are difficult to distinguish.  Additionally, the average accuracy of head poses is 100\%.

Table~\ref{tab:comp_KDEF} compares the accuracies on KDEF dataset using  our method, DenseNet~\cite{huang2017densely}, Transfer learning based CNN (TLCNN)~\cite{zhou2017action}, SURF boosting~\cite{rao2015multi},  and SVM~\cite{Moore11}.  Compared to TLCNN~\cite{zhou2017action} using a pre-training model, our method did not depend on any pre-trained CNN models. It achieved the accuracy of 88.5\% on FER and 100\% on pose estimation.  

\subsection{Experiments with AffectNet dataset}
\begin{figure}[!htb]
\centering
\centerline{\includegraphics[width=2.6in]{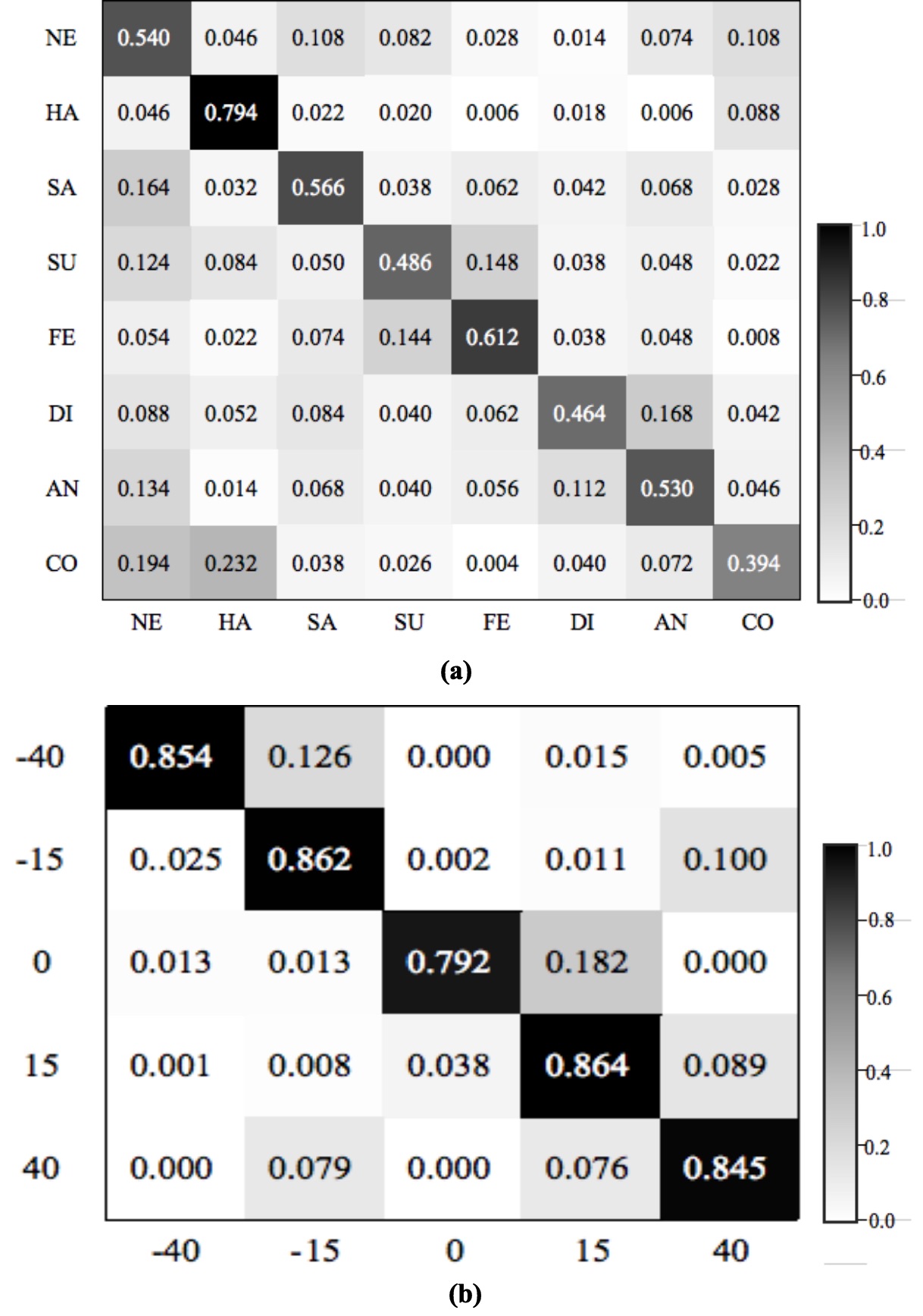}}
\caption{Overall performance on AffectNet dataset. (a)The confusion matrix for FER,
(b) the confusion matrix for head pose estimation.}\label{fig:AffectNet}
 \end{figure}
Fig. \ref{fig:AffectNet} shows the confusion matrix of FER and head poses. The average accuracy of FER achieved 54.82\%. The highest accuracy is 79.4\% of Happy followed by fear and sad, which achieve to 61.2\% and 56.6\%. The lowest accuracy is 39.4\% of Contempt. Relatively low accuracies appear between these expressions because the AffectNet dataset introduced more noises and spontaneous expressions than other datasets.

Table~\ref{tab:aff-p-F} shows the FER and pose accuracies using our method, DenseNet~\cite{huang2017densely}, VGG-19~\cite{Parkhi15}  and HOG+2 FC layers methods. It is shown that the accuracies of our method in multi-view FER are greater than the other methods under unconstraint environment.  
\begin{table}[H]
\centering
\caption{Comparison of accuracies (\%) using different methods on AffectNet. The highest accuracies are highlighted in bold.}
\resizebox{0.45\textwidth}{!}{%
\begin{tabular}{c|c|cc} 
\hline
Methods                                                       & Features  & Acc. on Poses   &  Acc. on FER  \\\hline\hline
\textbf{PhaNet  }                                      &{Pixel and scale attention} & \bf{85.3}& \bf{54.82} \\
DenseNet~\cite{huang2017densely}          & Facial image & 80.53 & 50.47\\ 
VGG-19~\cite{Parkhi15}                  & Facial image   & 77.85 & 32.90 \\
2 FC layers                                              & HOG   & 78.15 & 35.46  \\

\hline

\end{tabular}}
\label {tab:aff-p-F}
\end{table}

\subsection{Ablation study and discussion}

\subsubsection{Effect of different types of attention in PhaNet }

We further evaluated the effect of each individual attention component in our PhaNet model: no attention ($N_{S_1}$), hierarchical pixel attention at the second scale $H_{S_2}$ and third scale ($H_{S_3}$), joint scale attention in multi-scale representation part ($J_{SA}$). Table~\ref{tab:paths} shows that: (1) No attention ($N_{S_1}$) and any of pixel attention in each scale in isolation brings multi-view FER; (2) The combination of three $N_{S1}$, $H_{S2}$ and $H_{S3}$ gives further accuracy boost, which suggests the complementary information between the different regions discovered by our model; (3) When combining the pixel attention and scale attention, another significant performance gain is obtained. 

The recognition accuracies on four multi-view challenging datasets are summarized. 
Combining with $N_{S_1}$ and $H_{S_2}$, we can see that its performance is better than the result at a single scale. We obtain the second highest recognition accuracies by leveraging the power of pixel attention ensemble. 
Thanks to joint pixel attention and scale attention, the highest results are achieved on each dataset. It show that the proposed jointly hierarchical pixel attention and scale attention learning are capable of improving the performance of multi-view FER.


 \subsubsection{Visualization of expression-relevant region discovery}
 \begin{figure}[!htb]
 \centering
  \centerline{\includegraphics[width=3.4in]{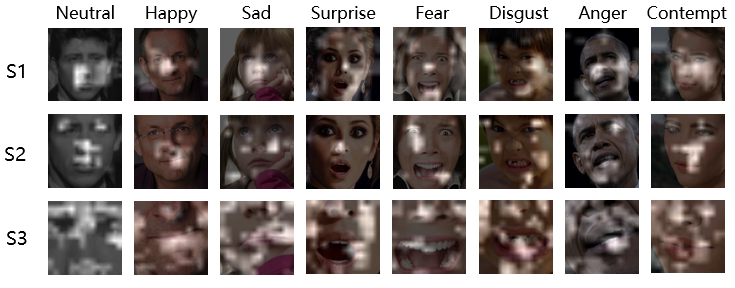}}
 \caption{ Attention maps of several test images in three hierarchical scales on AffectNet dataset. A highlight white denotes high attention with different scales. Better viewed in color and zoom in.
 }\label{fig:heat}
  \end{figure}
  
 \begin{figure} [!htb]
 \centering
  \centerline{\includegraphics[width=3.4in]{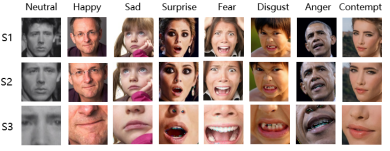}}
 \caption{ Examples of the discovered expression-relevant regions in 3 hierarchical scales under different expressions and poses on AffectNet dataset.  }\label{fig:regionsaff}
  \end{figure}
   
Fig. \ref{fig:heat} and Fig. \ref{fig:regionsaff} respectively show the learnt attention maps and discovered expression-relevant regions in three hierarchical scales on AffectNet dataset. It is obvious that two weakly-supervised PANs success to focus on different facial regions under different expressions, e.g., the mouth regions relevant to happy and left eye regions relevant to Neutral.

  \begin{figure}[!htb]
 \centering
  \centerline{\includegraphics[width=3.2in]{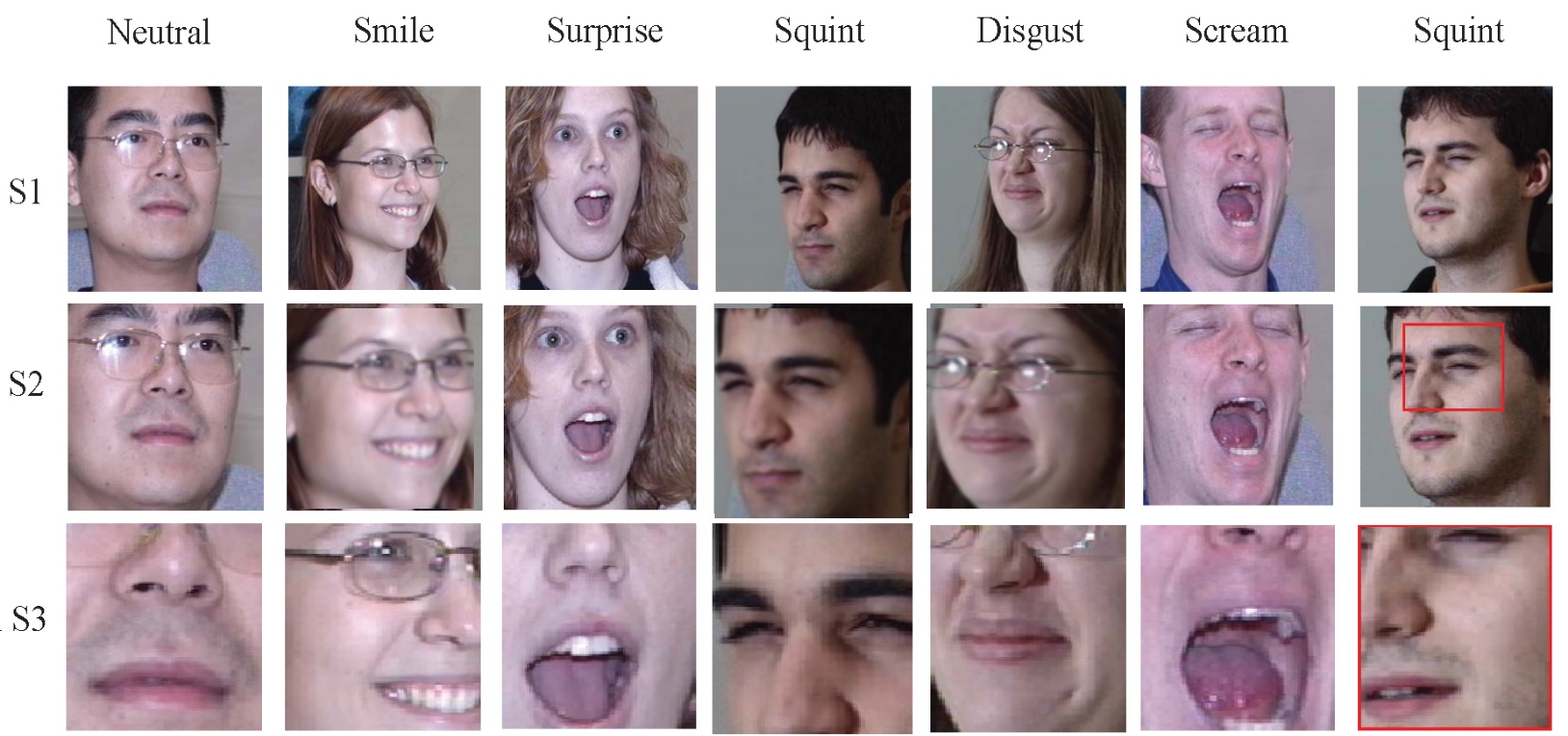}}
 \caption{ Examples of the discovered expression regions in 3 hierarchical scales under different expressions and poses on multi-view Multi-pie dataset.  }\label{fig:regions}
  \end{figure}

\begin{figure}[!htb]
 \centering
  \centerline{\includegraphics[width=3.0in, height=1.3in]{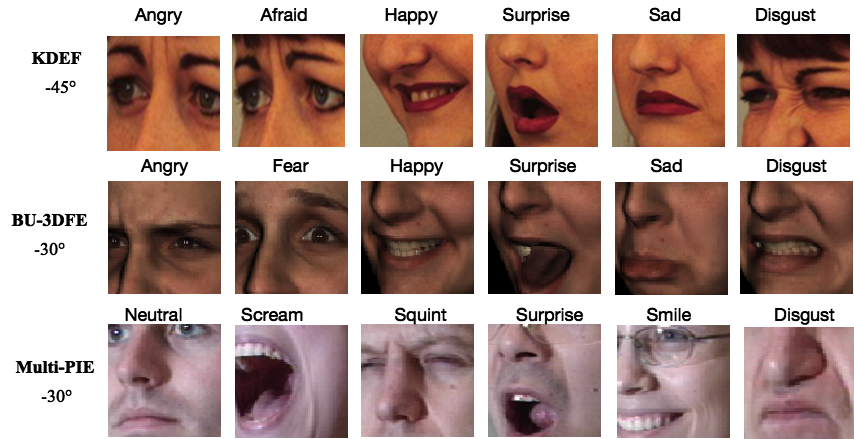}}
 \caption{ Examples of the discovered regions to different expressions with fixed poses on three datasets at the finer scale $S3$. From top to bottom: KDEF, BU-3DEF, Multi-pie.
 }\label{fig:region-s3}
  \end{figure}
  
Furthermore, Fig. \ref{fig:regions} show examples of the discovered expression-relevant regions in 3 hierarchical scales on multi-view Multi-pie dataset. PhaNet shows its capability in discovering expression-relevant regions under multi-view and unconstraint environment. We can see that these localized regions at scales $S_2$ and $S_3$ are consistent with human perception. These regions are clear and significant visual cues for FER.

\begin{figure}[!htb]
\centering
\centerline{\includegraphics[width=3in]{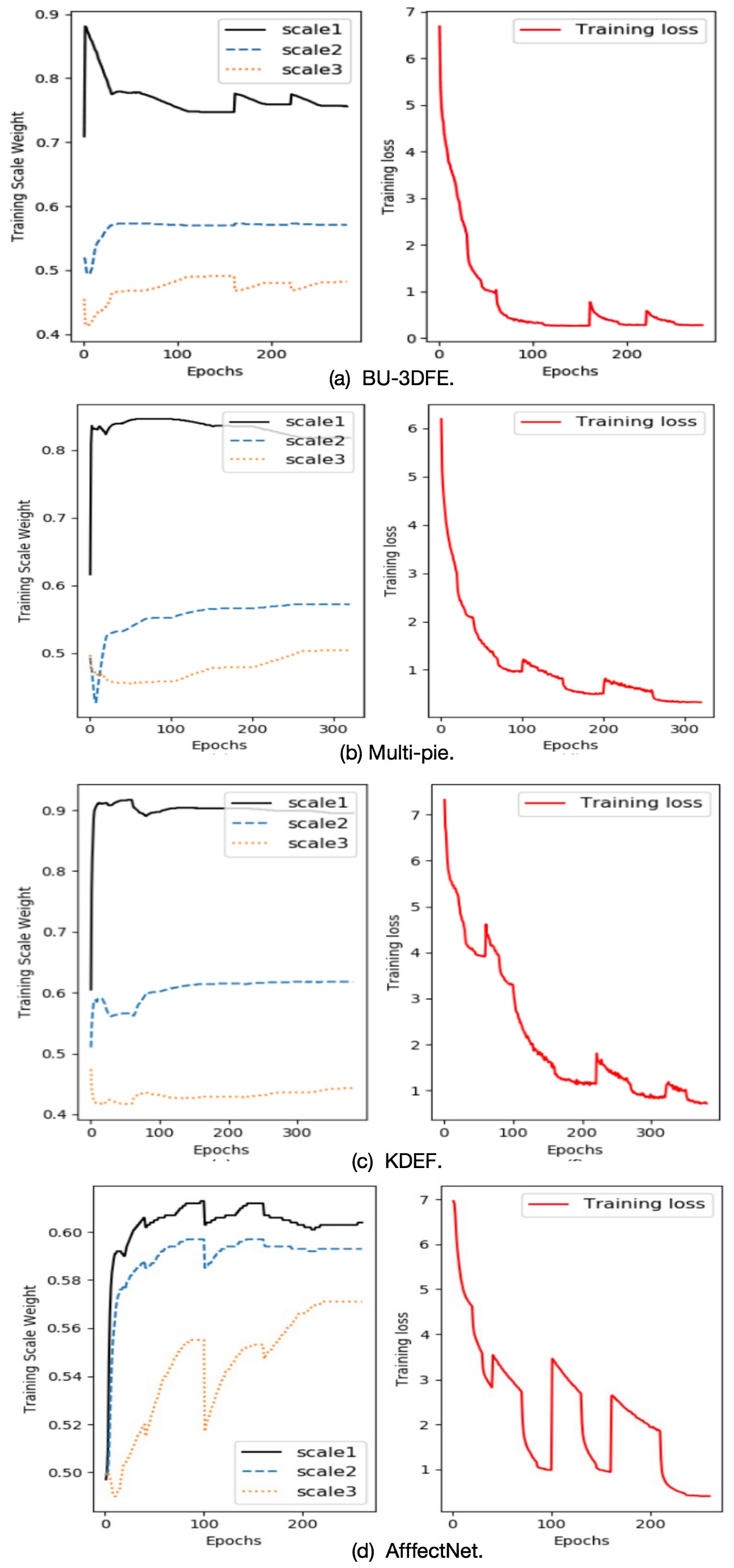}}
\caption{The joint SAL for PhaNet during the training process. Scale attention weights and loss with training epochs on the four datasets. From top to bottom: (a) BU-3DFE, (b) Multi-pie, (c) on KDEF, (d) AffectNet. }\label{fig:scale}
 \end{figure}
 
 \begin{figure*}
 \centering
  \centerline{\includegraphics[width=6.8in]{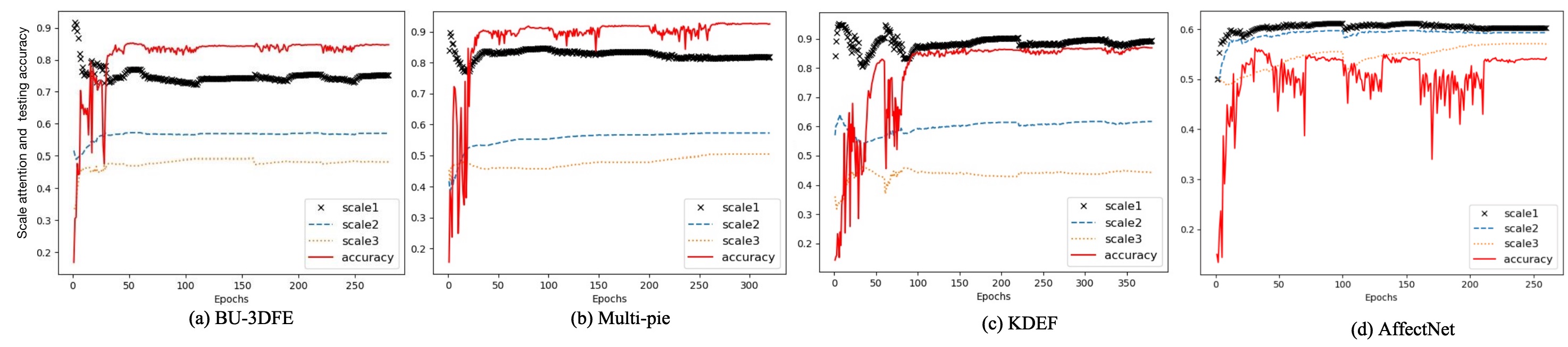}}
 \caption{  Learnt scale attention weights and FER accuracies with epochs during the prediction process on the four datasets. From left to right: (a) BU-3DFE, (b) Multi-pie, (c) KDEF, (d) AffectNet. }\label{fig:channel-testing}
  \end{figure*}

Moreover, to analysis the influence of the expressions at the finer scale, Fig. \ref{fig:region-s3} shows the finer expression-relevant regions at the scale $S_3$ with fixed poses on three multi-view datasets.  
From the results, we can observe that Happy/Smile, Surprise, and Scream expressions are relevant to mouth regions, while Squint, Disgust and Neutral expressions are more relevant to both eyes and mouth movement, while only angry expression is more relevant to eyes movement. Our method can give a guidance to localize the finer expression-relevant regions.

\subsubsection{Analysis on joint scale attention learning (SAL)}

In order to analysis quantitatively the learning procedure of joint SAL, Fig. \ref{fig:scale} provides the learned three scale attention weights and their training loss with epochs during the training process on the four datasets. Three scale attention weights have been initialized as the same value 0.5 in the first interactive step because they have been considered as the same importance for learning the representations by the Sigmoid function. As training goes on, the attention weight value of the first scale (see black solid lines) keeps growing then declining until convergence in 150 epochs, while the weight values of the second and third scale (see blue and yellow dotted lines) declines firstly then increasing latter tendency. Until the training has achieved convergence, the three scale attention weights are stabilized at different value respectively. As we expected, joint SAL is capable of learning a low weight for a pose-variant region and a high weight for a pose-invariant and informative one.

For evaluation scale attention on test procedure, Fig. \ref{fig:channel-testing} shows the accuracies of FER and attention weights with epochs during prediction on the four datasets. The average accuracies achieves all below 0.5 due to the unsuitable scale attention weights with less epochs. As the epoch goes on about 80, the average accuracies gradually increase on testing sets thanks to scale attention learning. 

\begin{figure}[!htb]
 \centering
  \centerline{\includegraphics[width=3.6in]{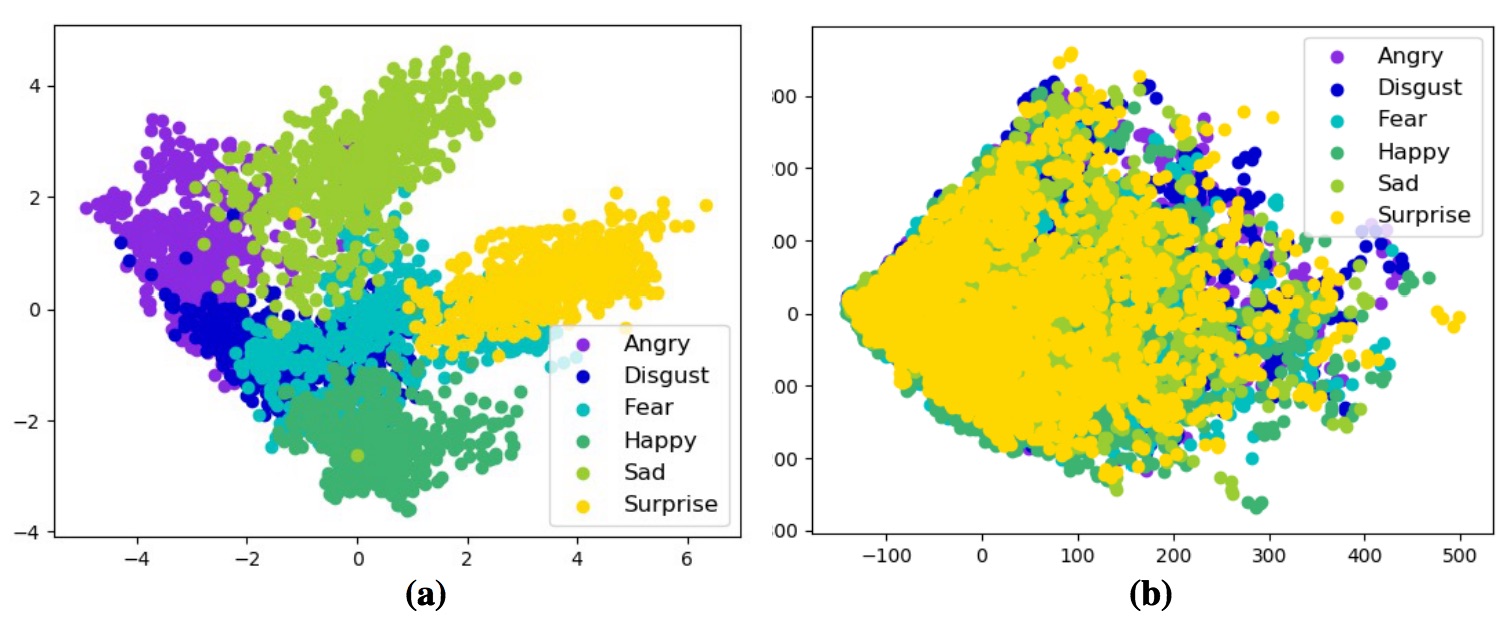}}
 \caption{ The comparison of different representations by t-SNE feature visualization. From left to right: (a) The joint pose-invariant and expression-discriminative representations in PhaNet by joint SAL. (b) the directly fused representations without using joint SAL. }\label{fig:channel-dis}
  \end{figure}
Moreover, to compare the features' discriminative power, we embed the joint multi-scale representations in PhaNet with joint SAL vs. the directly fused representations without using joint SAL to 2-D space by t-SNE ~\cite{hinton2008visualizing}, as shown in Fig. \ref{fig:channel-dis}. One can see that joint SAL can learn the more expression-discriminative and pose-invariant representations for multi-view FER.

\subsubsection{ Ablation analysis on the adopted dynamically weighting multi-loss in DCML}
To evaluate the impact of the adopted multi-loss in DCML, we compare ablation results under different loss settings by the proposed PhaNet for FER and pose accuracies on BU-3DFE dataset. All compared methods use the DenseNet backbone for part-based convolutional learning, but with different loss settings. 
The settings include: 1) optimizing parts only by minimizing FER loss $L(e)$,  2) only hierarchical learning by minimizing pixel attention loss $L_s^{att}$ and FER loss $L(e)$, (3) multi-task learning by minimizing the FER loss $L(e)$ and pose loss $L(p)$, 4) joint multi-task learning by minimizing the multi-loss with fixed weights $L_s^{att}+L(e)+L(p)$, 5) DCML by minimizing the dynamically weighting multi-losses $L_{multi}$ with a weighting constraint factor as Eq.(10). 
From the Table~\ref{tab:multi-losses}, 
the highest results for multi-view FER are achieved by minimizing the adopted $L_{multi}$ in the last row. As we expected, the pose loss helps reduce the impact of pose motion and the hierarchical attention loss helps learn the expression-discriminative representations.

\begin{table}[!htb]
\centering
\caption{Comparison of loss functions in terms of expression and pose classification accuracies on BU-3DFE dataset. }
\resizebox{0.45\textwidth}{!}{%
\begin{tabular}{c|cc}
\hline
Different loss                             &Acc. on   Pose    & Acc. on Exp.\\
\hline\hline
$L(e) $                                           &  -  & 77.79  \\
$L_s^{att}+L(e) $                            &  -  & 80.15  \\
$L(p)+L(e)$                                   & 98.25   & 79.34 \\
$L_s^{att}+L(e)+L(p) $                  &  98.70  & 82.27 \\
\hline
$L_s^{att}+{\lambda _{\rm{e}}}L(e)+{\lambda _{\rm{p}}}L(p)+{{e^{ - 12{\lambda _{\rm{p}}}(1 - {\lambda _{\rm{p}}})}}}$  & \bf{99.49}& \bf{84.92} \\
\hline
 \end{tabular}}
\label {tab:multi-losses}
\end{table}

\begin{figure}[!htb]
\centering
\centerline{\includegraphics[width=3.8in]{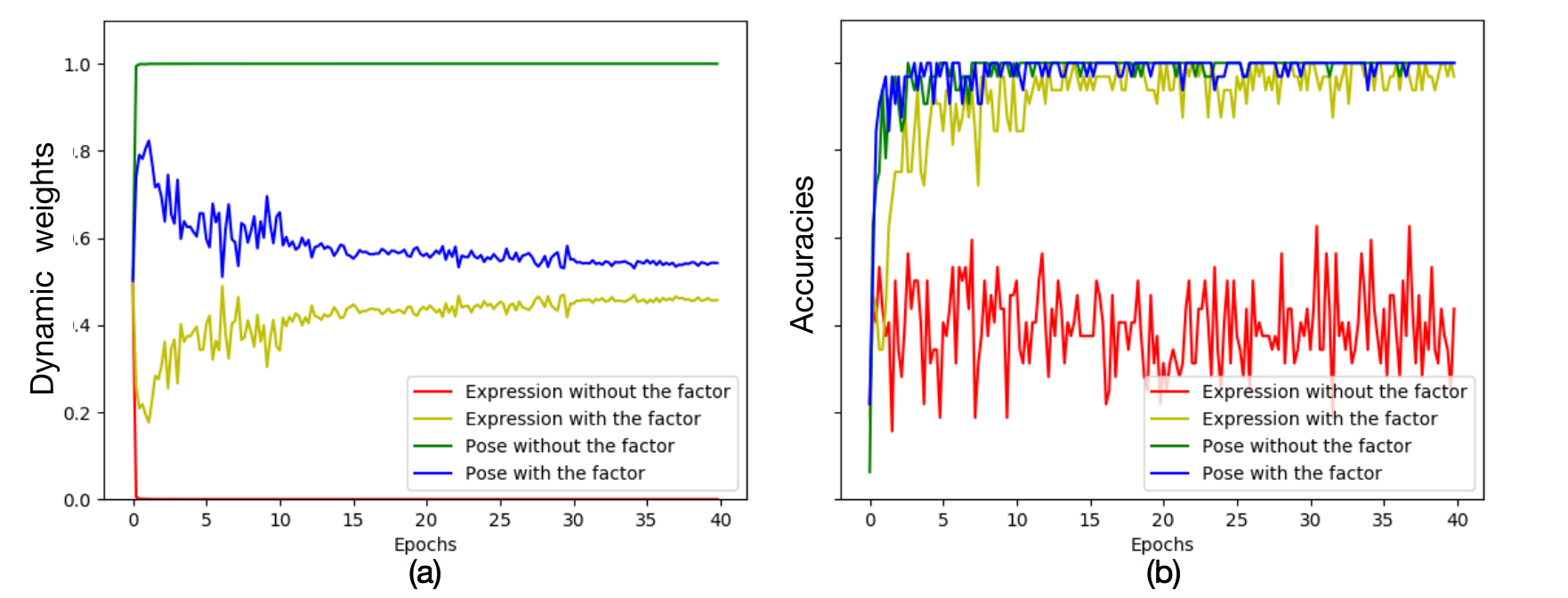}}
\caption{The learning procedure of dynamic weights and accuracies of each task for PhaNet model. (a) Dynamic weight learning with the factor vs. without the factor. (b) Pose and expression's accuracies with the factor vs. without the factor. }\label{fig:weightacc}
 \end{figure}

For further evaluating the dynamic weights and the weighting constraint factor in the adopted multi-loss, Fig.~\ref{fig:weightacc} shows the learning procedure of dynamic weights and accuracies of two tasks with vs. without using the constraint factor, respectively.
 
 With the constraint factor, the pose task has the larger weight in the beginning of training because it has the higher chance to be correct with random guess (see the blue curve in Fig.~\ref{fig:weightacc} (a)). As training goes on, the expression weight increases while the pose weight declines because of the constraint factor' influence (see the yellow curve in Fig.~\ref{fig:weightacc} (a)). When the learning converges, the dynamic weights of FER and poses reach about 0.6 and 0.4, and the accuracy of the two tasks reaches about 0.95 and 1 respectively (see the blue and yellow curves in Fig.~\ref{fig:weightacc} (b)). As we expected, dynamic weight learning promotes two tasks based on the contribution of each task, while the weighted constraint factor makes the whole network easier to converge.
 
Without the constraint factor,  ${\lambda _{\rm{e}}}$ rapidly decrease to 0 at the $1^{th}$ epoch due to the expression's gradient disappearance, and the accuracy of expression task declines about 0.4 (see the red curves in Fig.~\ref{fig:weightacc} (a)(b)). As we expected, the designed constrained factor is a symmetrical function whose domain is in $(0, 1] $. It can help to effectively avoid the gradient disappearance in the DCML procedure.

\subsection{Dataset crossing analysis}

To verify the generalization of the proposed PhaNet method, cross-dataset
experiments were carried out on very challenging in-the-wild dataset, as shown in Table~\ref{tab:crossing}. We used the trained BU-3DFE model to evaluate the very challenging SFEW dataset for facial expression prediction, because SFEW has no specific annotations for the head poses. 
Because training and testing dataset have absolutely different settings (e.g., pose, lighting, ethnicity,
glasses, age, etc.), the cross dataset task is much more difficult. However, compared to the state of the arts, our model also shows that it can be reusable for expression recognition on another dataset. 
The PhaNet achieved  31.25\% accuracy on the SFEW 
dataset; it all outperformed the GAN ~\cite{zhang2018joint}  and  DenseNet~\cite{huang2017densely}.  
The improvement mainly derives from the accurate attention localization, even though the cross-dataset SFEW without training.

\begin{table}[!htb]
\centering
\caption{Average accuracy (\%) of FER on across datasets. There FER models are trained on the BU-3DFE  dataset while tested on SFEW. }
\resizebox{0.45\textwidth}{!}{%
\begin{tabular}{|c|c|c|c|c|}
\hline
Methods & \textbf{PhaNet }   & GAN ~\cite{zhang2018joint}   & DenseNet~\cite{huang2017densely}  \\
\hline\hline
Accuracy & \bf{31.25} & 26.58  & 26.16 \\
\hline
\end{tabular}}
\label {tab:crossing}
\end{table}

 
\section{CONCLUSIONS  AND FUTURE WORKS}\label{sec:Conc}

This paper proposes an end-to-end trainable pose-adaptive hierarchical attention network (PhaNet) to jointly recognize the facial expressions and poses.
PhaNet discovers the most relevant regions to the facial expression by an attention mechanism in hierarchical scales, and then it selects the most pose-invariant scales to learn the pose-invariant and expression-discriminative representations. 
Experiments were conducted using four multi-view datasets and one crossing dataset. Thanks to jointly hierarchical pixel and scale attention learning and a dynamical weighting multi-task learning scheme, the proposed method achieved much improved performance and great robustness, with the highest accuracy of 93.53\% in multi-view FER and 100\% in pose estimation.  In future work, we will  introduce the hierarchical attention mechanism to model key information extraction in complex facial video sequences and analysis quantitatively for the finer expression-relevant regions to different expressions.

\section*{Acknowledgment}

The authors would like to thank Pro. Dacheng Tao for discussions,  Xunguang Wang for experimental analysis, and Dr. Xi Gong for chart drawing.

\bibliographystyle{IEEEtran}
\bibliography{ref}
%





%




\end{document}